\pdfoutput=1

\documentclass[11pt]{article}

\usepackage{ACL2023}

\usepackage{times}
\usepackage{latexsym}

\usepackage[T1]{fontenc}

\usepackage[utf8]{inputenc}
\usepackage{tcolorbox}
\tcbuselibrary{breakable}

\usepackage{microtype}

\usepackage{inconsolata}
\usepackage{mymathdef}

\newcommand{\FT}[0]{\includegraphics[width=.018\textwidth]{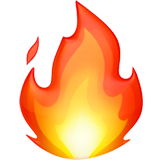}}
\newcommand{\FE}[0]{\includegraphics[width=.018\textwidth]{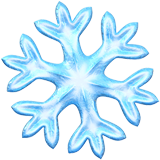}}

\title{SoftCoT: Soft Chain-of-Thought for Efficient Reasoning with LLMs}

\author{
Yige Xu${^{1,2}}$\thanks{\ \  The first two authors contributed equally.}, Xu Guo$^{1,*}$\footnotemark[2], Zhiwei Zeng$^1$\thanks{\ \  Corresponding authors.}, Chunyan Miao$^{1,2}$ \\
$^1$Joint NTU-UBC Research Centre of Excellence in Active Living for the Elderly \\
$^2$College of Computing and Data Science \\
Nanyang Technological University, Singapore \\
\texttt{\{yige002,xu008\}@e.ntu.edu.sg}, \texttt{\{zhiwei.zeng,ascymiao\}@ntu.edu.sg}
}

\begin{document}
\maketitle
\begin{abstract}
Chain-of-Thought (CoT) reasoning enables Large Language Models (LLMs) to solve complex reasoning tasks by generating intermediate reasoning steps. However, most existing approaches focus on hard token decoding, which constrains reasoning within the discrete vocabulary space and may not always be optimal. While recent efforts explore continuous-space reasoning, they often require full-model fine-tuning and suffer from catastrophic forgetting, limiting their applicability to state-of-the-art LLMs that already perform well in zero-shot settings with a proper instruction. To address this challenge, we propose a novel approach for continuous-space reasoning that does not require modifying the LLM. Specifically, we employ a lightweight fixed assistant model to speculatively generate instance-specific soft thought tokens as the initial chain of thoughts, which are then mapped into the LLM's representation space via a trainable projection module. Experimental results on five reasoning benchmarks demonstrate that our method enhances LLM reasoning performance through supervised, parameter-efficient fine-tuning. Source code is available at \url{https://github.com/xuyige/SoftCoT}.
\end{abstract}

\section{Introduction}

In recent years, Large Language Models (LLMs) have become a cornerstone in Natural Language Processing (NLP), exhibiting advanced natural language understanding and generation~\cite{DBLP:conf/nips/BrownMRSKDNSSAA20,du-etal-2022-glm,chowdhery2023palm,openai2023gpt4,touvron2023llama2,dubey2024llama,DBLP:journals/corr/abs-2412-15115}. Scaling model sizes has not only improved instruction-following~\cite{DBLP:conf/nips/KojimaGRMI22} but also triggered emergent reasoning abilities, as first evidenced by chain-of-thought (CoT) prompting~\cite{DBLP:conf/nips/Wei0SBIXCLZ22}. CoT prompts LLMs to generate intermediate reasoning steps before providing the final answer, which not only enhances interpretability but also improves a range of reasoning-intensive tasks~\cite{DBLP:conf/iclr/0001Z0S23,DBLP:journals/corr/abs-2409-12183}. It has inspired many advanced prompting frameworks, marking a paradigm shift from scaling training-time compute~\cite{DBLP:conf/nips/KojimaGRMI22} to scaling inference-time compute~\cite{DBLP:conf/iclr/0002WSLCNCZ23,DBLP:conf/nips/YaoYZS00N23} to further boost LLM performance.

Nevertheless, CoT's effectiveness depends on the quality of intermediate thoughts, as the auto-regressive generation process can propagate errors. To mitigate this challenge, methods like self-consistency~\cite{DBLP:conf/iclr/0002WSLCNCZ23} generate multiple reasoning paths, while Tree-of-Thought~\cite{DBLP:conf/nips/YaoYZS00N23} and Graph-of-Thought~\cite{DBLP:conf/aaai/BestaBKGPGGLNNH24} frameworks organize these paths to select higher-quality steps. Despite these improvements, such methods are computationally inefficient due to the need for extensive thought sampling.

To enhance CoT efficiency, recent research explores skipping the decoding of hard tokens at intermediate steps. Methods like Compressed CoT \citep{DBLP:journals/corr/abs-2412-13171} and Coconut \citep{DBLP:journals/corr/abs-2412-06769} conduct reasoning in a continuous space by using latent representations instead of discrete token sequences. Their results have shown to be superior to long-sequence discrete reasoning chains using only a short length of continuous representation. Yet, these methods require full-model fine-tuning, which incurs substantial computational costs, risks catastrophic forgetting, and limits their transferability across tasks.

We empirically observed that fine-tuning LLaMA3.1-8B~\cite{dubey2024llama} for continuous-space reasoning using a language modeling objective (as employed by Coconut and CCoT) results in performance degradation compared to zero-shot CoT (Tables~\ref{table:softcot-result-comparison}~and~\ref{table:softcot-generation-to-qwen2.5}). Drawing on a widely accepted definition of catastrophic forgetting~\cite{DBLP:journals/corr/abs-2401-05605,DBLP:journals/corr/abs-2411-15382}, defined as {\it the degradation of previously learned capabilities after fine-tuning on new data}, we conjecture that this drop in reasoning performance is attributable to {\bf catastrophic forgetting}. This phenomenon appears particularly pronounced in already capable instruction-tuned models such as LLaMA-3.1-8B-Instruct and Qwen2.5-7B-Instruct, which exhibit strong zero-shot CoT reasoning abilities. Thus, the methodologies of Coconut, which is based on GPT-2~\cite{radford2019language}, may not be directly applicable to more recent models such as LLaMA3.1 and Qwen2.5 series. Therefore, it is crucial to explore alternative methodologies that mitigate catastrophic forgetting while effectively leveraging continuous reasoning techniques in large-scale, instruction-tuned models, which is the main research goal of this work. To the best of our knowledge, we are the first to systematically identify and address the forgetting issue.

To mitigate catastrophic forgetting, a straightforward approach is to freeze the backbone LLM and instead optimize an external model for reasoning. Inspired by prompt tuning~\cite{DBLP:conf/emnlp/LesterAC21} and speculative decoding \citep{DBLP:conf/icml/LeviathanKM23}, we propose to utilize an auxiliary small assistant model to generate a sequence of ``thought'' tokens conditioned on a task instruction followed by a specific instance~\cite{DBLP:conf/nips/LiPHGGY23,DBLP:conf/icml/ShaoGSHDC23}. These tokens serve as instance-specific prompts that  adapt to different problems to boost LLM's reasoning. Such an auxiliary prompting mechanism allows the LLM to achieve better generalization while preserving its pre-trained knowledge.

\begin{figure*}[!thp]
  \centering
  \includegraphics[width=\textwidth]{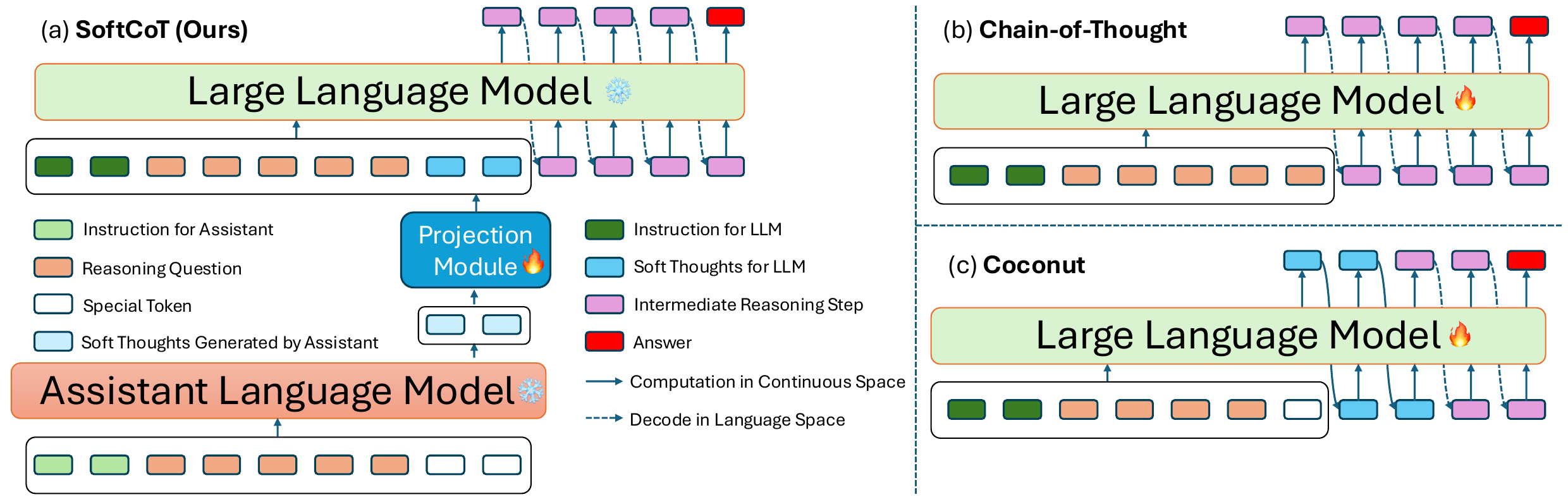}
\caption{A comparison of SoftCoT, vanilla Chain-of-Thought, and Coconut.}\label{fig:softcot-methodology-overview}
\end{figure*}

To exploit continuous-space reasoning, we use the last-layer hidden states from the small assistant model as the ``soft'' thought tokens, rather than the discrete tokens obtained after vocabulary mapping. Staying in the latent space avoids information loss inherent in autoregressive decoding. However, a representational gap between the assistant model and the LLM may hinder effective knowledge transfer. To bridge this gap, we train a projection module to map the soft thought tokens generated by the assistant model to the LLM's representation space. Training the projection module for each task can be seen as \textit{soft prompt tuning} for the LLM. The overall \underline{\bf Soft} thoughts for \underline{\bf CoT} (SoftCoT) reasoning framework is illustrated in Figure~\ref{fig:softcot-methodology-overview}.

We evaluate SoftCoT on five reasoning benchmarks and two state-of-the-art LLM architectures. The five benchmarks include mathematical reasoning, commonsense reasoning, and symbolic reasoning. For further exploration, we create a hard version of the ASDiv dataset~\cite{DBLP:conf/acl/MiaoLS20}, which requires stronger mathematical reasoning ability. The new dataset is named ``ASDiv-Aug'' in this paper. Experimental results show that SoftCoT consistently improves accuracy on both public datasets and our augmented ASDiv-Aug dataset, demonstrating the effectiveness of SoftCoT in enhancing LLM's reasoning performance. Moreover, SoftCoT effectively mitigates catastrophic forgetting seen in previous methods based on full-model fine-tuning, and our results highlight the effectiveness of using an assistant model to generate soft thoughts. SoftCoT thus represents the first lightweight yet powerful approach that leverages the benefits of soft thought tokens while preserving LLM' prior knowledge.

\section{Related Works}

Early research on chain-of-thought (CoT) reasoning can be traced back to \citet{DBLP:conf/nips/Wei0SBIXCLZ22}, who first introduced a prompting strategy that guides LLMs through decomposed intermediate reasoning steps using few-shot exemplars. Concurrently, \citet{DBLP:conf/nips/KojimaGRMI22} demonstrated that LLMs are capable of zero-shot CoT reasoning by simply appending the phrase ``Let's think step by step'' to the prompt template. This discovery underscored the latent reasoning abilities of LLMs, even in the absence of explicit demonstrations.

Building upon these foundational works, the NLP community has extensively explored the potential of CoT reasoning. As summarized by \citet{DBLP:conf/acl/ChuCCYH0P00L24}, recent advancements in CoT methods can be broadly categorized into three areas: (1) {\it Prompt Construction}, which aims to optimize prompts for improved CoT reasoning \cite{DBLP:conf/nips/Wei0SBIXCLZ22,DBLP:conf/nips/KojimaGRMI22,DBLP:conf/iclr/0001Z0S23}; (2) {\it Topological Variants}, which leverage structured representations such as trees and graphs to enhance CoT reasoning~\cite{DBLP:conf/nips/YaoYZS00N23,DBLP:conf/aaai/BestaBKGPGGLNNH24}; and (3) {\it Enhancement Methods}, which introduce external strategies to further improve CoT reasoning, such as question decomposition~\cite{DBLP:conf/iclr/ZhouSHWS0SCBLC23} and self-consistency decoding \cite{DBLP:conf/iclr/0002WSLCNCZ23}. Despite the effectiveness of these approaches, the majority of existing CoT methods rely on discrete token-by-token generation, which imposes inherent constraints and limits their expressiveness.

To address the limitations of discrete language space, an effective approach is to leverage continuous representation space for reasoning. Coconut~\cite{DBLP:journals/corr/abs-2412-06769} introduces a Chain-of-Continuous-Thought, while CCoT~\cite{DBLP:journals/corr/abs-2412-13171} employs Compressed Chain-of-Thought, generating content-rich and continuous contemplation tokens. Heima~\cite{DBLP:journals/corr/abs-2501-19201} further advances this idea by utilizing a single continuous vector to represent compressed reasoning tokens in multi-modal tasks. However, both Coconut and CCoT rely on a language modeling objective for supervised fine-tuning, which is infeasible for state-of-the-art LLMs due to the catastrophic forgetting problem. Moreover, Heima underperforms compared to its backbone model, LLaVA-CoT~\cite{DBLP:journals/corr/abs-2411-10440}. These challenges underscore the need to develop methodologies that mitigate catastrophic forgetting in the application of continuous-space CoT reasoning.

\section{Methodology}

\subsection{Problem Definition and Notations}
\label{sec:softcot-methodology-problem-def}

Given a question $\cQ=[q_1,q_2,\cdots,q_{|\cQ|}]$, the CoT reasoning will solve the problem on the following two steps: (1) Auto-regressively generate a list of rationale steps $\cR=[r_1,r_2,\cdots,r_{|\cR|}]$ according to the question; (2) Auto-regressively generate the answer $\cA=[a_1,a_2,\cdots,a_{|\cA|}]$ according to the question as well as the rationale steps. The generation process can be described as:
\begin{align}
  r_{i+1}&=\mathrm{LLM}(\cQ;\cR_{\le i}),\\\nonumber
  a_{j+1}&=\mathrm{LLM}(\cQ;\cR;\cA_{\le j}),
\end{align}
where $\mathrm{LLM}(\cdot)$ indicates a large language model, $\cR_{\le i}=[r_1,\cdots,r_i]$ indicates the previous generated $i$ reasoning tokens, and $\cA_{\le j}=[a_1,\cdots,a_j]$ indicates the previous generated $j$ answer tokens.

The majority of recent works~\cite{DBLP:conf/iclr/0001Z0S23,DBLP:conf/iclr/ZhouSHWS0SCBLC23,DBLP:conf/nips/YaoYZS00N23} focus on generating discrete hard tokens in $\cR$, which is named as ``{\bf Hard-CoT}'' in this paper. On contrast, some recent works~\cite{DBLP:journals/corr/abs-2412-06769,DBLP:journals/corr/abs-2412-13171} focus on the continuous representations ({\it a.k.a} latent space) of $\cR$, which is named as ``{\bf Soft-CoT}'' in this paper.

In this paper, we mannually define some rules (e.g., regular expression matching) to extract the framework answer $\hat{\cA}$ from the answer $\bar{\cA}$ generated by the LLM. Then we compute the accuracy of $\hat{\cA}$ comparing with the ground-truth answer $\cA$.

\subsection{Overview of the SoftCoT Framework}

SoftCoT is a novel framework designed to enhance reasoning capabilities in large language models (LLMs). Given an input question $\cQ$, the framework produces both a sequence of reasoning steps $\cR$ and the final answer $\cA$. SoftCoT consists of three key components: the soft thought token generation module, the projection module, and the CoT reasoning module. The overall architecture is illustrated in Figure~\ref{fig:softcot-methodology-overview}(a).

The soft thought token generation module is inspired by prompt tuning techniques~\cite{DBLP:conf/emnlp/LesterAC21}. In conventional prompt tuning, learnable prompts facilitate the retrieval of knowledge stored within the LLM~\cite{xu-etal-2023-efficient}. In SoftCoT, soft thought tokens are generated by an assistant language model, which is typically smaller than the backbone LLM (e.g., LLaMA-3.2-1B-Instruct as the assistant model and LLaMA-3.1-8B-Instruct as the backbone model).

A key challenge in this setup is that the assistant model can only generate discrete token sequences as input to the backbone LLM, which imposes constraints and may not always yield optimal prompts. To address this limitation, we introduce continuous soft thought tokens that enable more expressive and flexible prompting. However, a representation gap exists between the assistant model and the backbone LLM, necessitating an intermediate transformation.

To bridge this gap, the projection module maps the soft thought tokens' representations into a space more compatible with the backbone LLM. This ensures that the soft thought tokens effectively guide the reasoning process.

Finally, the CoT reasoning module leverages both the learned soft thought tokens and word embeddings to generate intermediate reasoning steps $\bar{\cR}$ and the final answer $\bar{\cA}$. The model is trained using a language modeling objective, optimizing the learnable parameters across the rationale steps and the answer spans.

\subsection{Prompt Tuning for CoT Reasoning}

Prompt tuning for CoT reasoning aims to optimize the structure and content of the prompt template to enhance the reasoning capabilities of a large language model (LLM). This process can be mathematically formulated as follows:
\begin{align}
  \hat{y}&=\mathrm{LLM}\big(P_{\bp}(x)\big),\\\nonumber
  \bp^*&=\arg \min_{\bp} \cL(\hat{y},y),
\end{align}
where $\hat{y}$ represents the predicted output, $x$ denotes the input sequence, and $P_{\bp}(x)$ is the input augmented with a prompt $\bp$. The objective function $\cL(\cdot)$ measures the discrepancy between the model's prediction $\hat{y}$ and the ground-truth label $y$. The primary goal of prompt tuning is to determine an optimal prompt configuration that effectively guides the LLM to perform CoT reasoning with improved accuracy and interpretability.

A straightforward yet effective approach to optimizing prompts involves leveraging an auxiliary assistant model to generate instance-specific prompts, which provide contextual hints or question summaries to facilitate reasoning~\cite{DBLP:conf/nips/LiPHGGY23,DBLP:conf/icml/ShaoGSHDC23,li2024towards}. In this framework, the prompt $\bp$ can be decomposed into two components: (1) a fixed, task-specific prompt $\bp_{\FE}$,  which remains constant across all instances and encodes general problem-solving heuristics, and (2) a learnable, instance-specific prompt $\bp_{\FT}$, which dynamically adapts to each input instance to provide tailored guidance.

Given the rapid advancements in LLMs, many LLMs are capable of solving complex reasoning tasks under zero-shot settings. Instead of fine-tuning the assistant model for each task, we adopt a more efficient strategy by employing a relatively small, frozen language model to generate $\bp_{\FT}$. This approach not only reduces computational costs but also ensures stability and generalization across different problem domains. By systematically integrating instance-specific prompting with fixed task-specific instructions, this method enhances the LLM's reasoning process while maintaining adaptability to various contexts.

\subsection{Soft Thought Tokens for CoT Reasoning}

One of the advantages of Hard-CoT is that the generated discrete tokens can be tokenized by the LLMs, which does not require a external mapping module. However, there are two main limitations of Hard-CoT: (1) The decoded token space is discrete, which is constrained and sometimes not optimal; (2) The gradient cannot backpropagate to the assistant model since the decoding process cut off the gradient information. A convincing solution is replace the hard tokens to soft thought tokens.

\paragraph{Generating Soft Thought Tokens with an Assistant Model}

To generate instance-specific soft thoughts, we utilize an auxiliary assistant model that produces soft thoughts based on the given reasoning task. The input to the assistant model, denoted as $\bx_{\mathrm{assist}}$, consists of three main components:
\begin{align}
  \bx_{\mathrm{assist}}=\mathrm{concat}\big[\cI_{\mathrm{assist}},\cQ,\texttt{[UNK]}_{1:N}\big],\\\nonumber
\end{align}
where
\begin{itemize}
  \item $\cI_{\mathrm{assist}}$ represents the instructional context provided to the assistant model, guiding it on how to generate relevant thoughts.
  \item $\cQ$ denotes the reasoning question that the primary LLM will solve, which has been defined in Section~\ref{sec:softcot-methodology-problem-def}.
  \item $N$ \texttt{[UNK]} tokens serve as placeholders for the soft thoughts.
\end{itemize}

Once the input sequence is constructed, the assistant model processes it, and the soft thought tokens are obtained as follows:
\begin{align}
  \bh^{\mathrm{assist}}&=\mathrm{Assistant}(\bx_{\mathrm{assist}}),\\\nonumber
  \bt_{\mathrm{assist}}&=\bh^{\mathrm{assist}}_{|\cI|+|\cQ|+1:|\cI|+|\cQ|+N}.
\end{align}
Here $\bh^{\mathrm{assist}}$ denotes the final-layer hidden states of the assistant model, and $\bt_{\mathrm{assist}}$ corresponds to the segment of $\bh^{\mathrm{assist}}$ associated with the $N$ \texttt{[UNK]} tokens. This extracted representation serves as the instance-specific soft thoughts, dynamically adapting to the input reasoning question.

\paragraph{Projection Module}

Since there exist both a representation gap and a dimensional gap between the assistant language model and the primary LLM, a direct utilization of $\bt_{\mathrm{assist}}$ may lead to suboptimal performance. The assistant model and the LLM often operate in different embedding spaces, with distinct hidden state distributions and dimensionalities. To bridge this gap, we introduce a projection module that maps the assistant-generated soft thoughts $\bt_{\mathrm{assist}}$ from the assistant model's embedding space to the LLM's embedding space:
\begin{align}
  \cT_{\mathrm{soft}}=\mathrm{Linear}_{\theta}(\bt_{\mathrm{assist}}),\label{eq:softcot-projection}
\end{align}
where $\mathrm{Linear}_{\theta}:\mathbb{R}^{d_\mathrm{assist}}\to \mathbb{R}^{d_\mathrm{LLM}}$ is a trainable projection layer parameterized by $\theta$. This layer ensures that the assistant-generated soft thoughts are transformed into a suitable format for the LLM, preserving relevant semantic information while adapting to the LLM’s feature space.

By incorporating this projection module, we effectively mitigate discrepancies between the assistant model and the LLM, enabling smooth integration of instance-specific thoughts into the CoT reasoning process. This design ensures that the learned thought tokens are both informative and compatible, thereby enhancing the overall reasoning performance of the LLM.

\paragraph{LLM Reasoning with Soft CoT}
With instance-specific soft thought tokens generated by the assistant model and mapped to the LLM's embedding space, we proceed to the final step: applying these soft thoughts to aid LLMs in CoT reasoning.

The input to the LLM, denoted as $\bx_{\mathrm{LLM}}$, follows a structure similar to that of  $\bx_{\mathrm{assist}}$:
\begin{align}
   \bx_{\mathrm{LLM}}=\mathrm{concat}\big[\cI_{\mathrm{LLM}},\cQ,\cT_{\mathrm{soft}}\big],\label{eq:softcot-llm-iput-def}
\end{align}
where
\begin{itemize}
  \item $\cI_{\mathrm{LLM}}$ is the task-specific instruction, which is a fixed prompt shared across all instances of the same task. It provides general problem-solving heuristics and instructions relevant to the reasoning task.
  \item $\cT_{\mathrm{soft}}$ is the instance-specific soft thoughts computed by Eq~\eqref{eq:softcot-projection}. This component dynamically adapts soft thought tokens to each input question, enhancing contextual understanding.
\end{itemize}

With this structured input, the LLM generates step-by-step reasoning chains, following the principles of CoT reasoning. The reasoning process unfolds by systematically applying logical deductions or problem-solving heuristics, ultimately leading to the generation of the final answer:
\begin{align}
  \bar{\cR}&=\mathrm{LLM}(\bx_{\mathrm{LLM}}),\\\nonumber
  \bar{\cA}&=\mathrm{LLM}(\bx_{\mathrm{LLM}},\bar{\cR}),\\\nonumber
  \hat{\cA}&=\cE(\bar{\cA}),
\end{align}
where $\cE(\cdot)$ is mannual rules for answer extraction.

By integrating both fixed task-specific instructions and instance-specific soft thought tokens, our approach enables the LLM to systematically decompose complex reasoning tasks while leveraging auxiliary knowledge provided by the assistant model. The structured input ensures that the LLM benefits from both general domain knowledge and tailored instance-level guidance, ultimately improving its reasoning effectiveness.

\paragraph{Parameter-Efficient Training}

In this work, we focus on reasoning tasks that include annotated reasoning steps, which provide explicit intermediate reasoning trajectories leading to the final answer. To effectively train the model, we employ the standard language modeling objective (also known as next-token prediction) to supervise the generation of soft thoughts. During the training stage, the input sequence is structured as follows:
\begin{align}
   \bx_{\mathrm{train}}=\mathrm{concat}\big[\cI_{\mathrm{assist}},\cQ,\cT_{\mathrm{soft}},\cR,\cA\big].
\end{align}

To effectively learn the soft thoughts, we apply the negative log-likelihood (NLL) loss over the reasoning steps and the answer span. Specifically, we mask the tokens before the intermediate reasoning steps to prevent the model from directly relying on them during loss computation. Instead, the model is trained to generate the reasoning steps $\cR$ and final answer $\cA$ in an autoregressive manner.

\begin{table*}[t!]
    \centering\small
    \tabcolsep 6.0pt
    \begin{tabular}{l|c c c c }
    \toprule
    Dataset & Task Type & Answer Type & \# Train samples & \# Evaluation samples \\
    \midrule
    GSM8K & \multirow{3}{*}{Mathematical} & Number & 7,473 & 1,319\\
    ASDiv-Aug & ~ & Number & 4,183 & 1,038 \\
    AQuA & ~ & Option & 97,467 & 254 \\
    \midrule
    StrategyQA & Commonsense & Yes/No & 1,832 & 458 \\
    \midrule
    DU & Symbolic & Option & - & 250 \\
    \bottomrule
\end{tabular}
    \caption{Summary statistics of five datasets we used. ``-'' indicates that there is no training samples available. ``DU'' indicates the Date Understanding~\cite{DBLP:journals/tmlr/SrivastavaRRSAF23} dataset.}\label{table:softcot-experiment-dataset}
\end{table*}

\section{Experiments}

\subsection{Datasets}

We conduct experiments on five reasoning datasets spanning three categories of reasoning: mathematical reasoning, commonsense reasoning, and symbolic reasoning. For mathematical reasoning, we utilize {\bf GSM8K}~\cite{cobbe2021gsm8k}, {\bf ASDiv}~\cite{DBLP:conf/acl/MiaoLS20}, and {\bf AQuA}~\cite{DBLP:conf/acl/LingYDB17}. For commonsense reasoning, we employ {\bf StrategyQA}~\cite{DBLP:journals/tacl/GevaKSKRB21}. For symbolic reasoning, we use {\bf Date Understanding}~\cite{DBLP:journals/tmlr/SrivastavaRRSAF23} from the BIG-benchmark.

Given that LLaMA-3.1-8B-Instruct is a well-trained LLM, we augment the ASDiv dataset to ensure that the model encounters novel instances. Specifically, we replicate each instance five times and systematically extract and replace numerical values in the questions with randomly selected alternatives. For example, the original dataset is ``{\bf eat 6 apples per day}'', we will duplicate this instance multiple times and {\bf replace 6 with 7, or 8, or 12, or 18 randomly}. This augmentation is designed to evaluate the model's reasoning capability rather than its ability to recognize patterns from memorized data. The augmented dataset is named as ``{\bf ASDiv-Aug}'' in the following part of this paper. We open-source our ASDiv-Aug to the community at \url{https://huggingface.co/datasets/xuyige/ASDiv-Aug}. All detail statistics of the datasets is shown in Table~\ref{table:softcot-experiment-dataset}.

\begin{table*}[t!]
\setlength{\tabcolsep}{4pt}
    \centering
    \begin{tabular}{l| c c c | c | c | c c c c c}
    \toprule
    \multirow{2}{*}{Model} & GSM8K & ASDiv-Aug & AQuA & StrategyQA & DU & \multirow{2}{*}{Avg.} \\
    \cline{2-6}
    ~ & \multicolumn{3}{c|}{Mathematical} & Commonsense & Symbolic & ~ \\
    \midrule
    {\it GPT-2} & \\
    Coconut~\cite{DBLP:journals/corr/abs-2412-06769}
    & 34.10$^{*}_{\pm \text{1.50}}$ & 38.92$^{\dagger}_{\pm \text{0.00}}$ & 22.83$^{\dagger}_{\pm \text{0.00}}$ & - & - & - \\
    \midrule
    \midrule
    {\it LLaMA-3.1-8B-Instruct}\\
    Zero-Shot CoT
    & 79.61$_{\pm \text{0.81}}$ & 86.78$_{\pm \text{0.63}}$ & 54.65$_{\pm \text{2.43}}$ & 65.63$_{\pm \text{3.31}}$ & 54.40$_{\pm \text{2.40}}$ & 68.21 \\
    Zero-Shot CoT-Unk
    & 79.95$_{\pm \text{0.59}}$ & 86.90$_{\pm \text{0.41}}$ & 55.28$_{\pm \text{1.88}}$ & 66.16$_{\pm \text{2.70}}$ & 54.16$_{\pm \text{1.46}}$ & 68.49 \\
    Zero-Shot Assist-CoT
    & 80.76$_{\pm \text{1.53}}$ & 86.96$_{\pm \text{0.46}}$ & 55.83$_{\pm \text{2.98}}$ & 66.55$_{\pm \text{3.99}}$ & 58.24$_{\pm \text{3.56}}$ & 69.67 \\
    \midrule
    LoRA Fine-Tuning
    & 75.66$_{\pm \text{0.00}}$ & 86.67$_{\pm \text{0.00}}$ & 52.36$_{\pm \text{0.00}}$ & - & - & - \\
    Coconut~\cite{DBLP:journals/corr/abs-2412-06769}$^{\dagger}$
    & 76.12$_{\pm \text{0.00}}$ & 86.80$_{\pm \text{0.00}}$ & 53.15$_{\pm \text{0.00}}$ & - & - & - \\
    {\bf SoftCoT (Ours)}
    & {\bf 81.03$_{\pm \text{0.42}}$} & {\bf 87.19$_{\pm \text{0.40}}$} & {\bf 56.30$_{\pm \text{1.67}}$} & {\bf 69.04$_{\pm \text{1.23}}$} & {\bf 59.04$_{\pm \text{1.93}}$} & {\bf 70.52} \\
    \bottomrule
\end{tabular}
\caption{Model comparison with baselines. ``DU'' indicates the Date Understanding~\cite{DBLP:journals/tmlr/SrivastavaRRSAF23} dataset. The first row is under the backbone of GPT-2~\cite{radford2019language} as backbone. The following rows are under the backbone of LLaMA-3.1-8B-Instruct~\cite{dubey2024llama}. The last three rows are models trained via the language modeling objective. We run for 5 random seeds and report the average accuracy as well as the standard variance. ``*'' indicates that the accuracy is reported by \citet{DBLP:journals/corr/abs-2412-06769}. ``$^{\dagger}$'' indicates the results that we modify and run the official code of Coconut. ${\pm \text{0.00}}$ indicates that we only run once for baseline results.
\label{table:softcot-result-comparison}}
\end{table*}

\subsection{Baselines}
\label{sec:softcot-exp-baselines}

Considering that zero-shot CoT on modern LLMs already sets a strong baseline, often outperforming fine-tuned models due to their rich pretraining, we will consider zero-shot baseline as a fair comparison. Meanwhile, we also consider baselines the fine-tunes the LLM:

\paragraph{Zero-Shot CoT} We adopt the prompt templates from \citet{DBLP:journals/corr/abs-2409-12183} to test zero-shot CoT performance. This baseline serves to assess whether the model experiences performance degradation after supervised fine-tuning.

\paragraph{Zero-Shot CoT-Unk} We directly append some \texttt{[UNK]} tokens to represent the un-tuned prompts for the LLM to perform CoT reasoning. This baseline evaluates the effectiveness of projection tuning for soft thought tokens.

\paragraph{Zero-Shot Assist-CoT} The assistant model is prompted to generate a hard-token sequence under standard CoT prompting, truncated at 24 tokens. This sequence is then used as a prompt for the LLM to perform CoT reasoning. This baseline evaluates the effectiveness of soft thoughts by comparing them with hard-token prompts.

\paragraph{Coconut}~\citet{DBLP:journals/corr/abs-2412-06769} propose training LLMs to reason in a continuous latent space by iteratively feeding hidden states from the previous step as input embeddings to the next step. The continuous thought encodes rich information, allowing the model to explore more effective reasoning paths. We use their official code\footnote{\url{https://github.com/facebookresearch/coconut}} to implement this baseline. To adapt Coconut to larger Llama3.1 and Qwen2.5 models, we apply LoRA fine-tuning.

\paragraph{LoRA Fine-Tuning} We apply LoRA fine-tuning~\cite{DBLP:conf/iclr/HuSWALWWC22} ($r=16$) with the language modeling objective as our baseline. This baseline examines the effectiveness of appending soft thoughts to LLMs compared to traditional parameter-efficient methods like LoRA.

Implementation details for baselines as well as SoftCoT is shown in Appendix~\ref{sec:softcot-appendix-implementation-details}.

\section{Results and Discussions}

\subsection{Comparison with Baselines}

To evaluate SoftCoT, we compare its performance against the baselines introduced in Section~\ref{sec:softcot-exp-baselines}. The results are summarized in Table~\ref{table:softcot-result-comparison}:

\noindent (1) {\bf Supervised LoRA Fine-Tuning performs worse than zero-shot CoT, which make Coconut not applicable to larger language models}: We modify and run the official implementation of Coconut, adapting it to LLaMA-3.1-8B-Instruct. Our findings indicate that Coconut exhibits performance degradation following supervised fine-tuning with the language modeling objective. Further experiments on LoRA Fine-Tuning show that it fails to match the performance of zero-shot CoT, which demonstrates that the updated LLM after supervised fine-tuning will have a degradation of previously learned capabilities. We conjecture this drop in reasoning performance is attributable to {\bf catastrophic forgetting}, which aligns with findings from prior studies, including \citet{DBLP:journals/corr/abs-2401-05605} and \citet{DBLP:journals/corr/abs-2411-15382}, which have reported similar issues. Thus, to mitigate the catastrophic forgetting, we fixed the whole LLM in SoftCoT.

\noindent (2) {\bf Incorporating \texttt{[UNK]} tokens mitigates performance variance}: We examined the effect of directly adding \texttt{[UNK]} tokens as thoughts $\cT_{\mathrm{soft}}$ in Eq.~\eqref{eq:softcot-llm-iput-def}. The results demonstrate a slight improvement in overall performance and a reduction in variance. The \texttt{[UNK]} token, also known as the ``pause token''~\cite{DBLP:conf/iclr/GoyalJRMKN24}, appears to expand the model's computational capacity, leading to more stable and consistent outputs.

\noindent (3) {\bf Assistant model is effective to facilitate CoT reasoning}: We utilize instruction to require the assistant model generate some hard prompts, which can be regarded as the initial thoughts for CoT reasoning. Experiment results show that although it has a larger variance than CoT-Unk, it facilitates the LLM for more diverse CoT generation, which leads to the performance improvement from 68.49 to 69.67 in average.

\noindent (4) {\bf SoftCoT consistently benefits from the supervised fine-tuning}: Overall, our proposed SoftCoT consistently outperforms baselines across all five reasoning datasets, involving the mathematical reasoning, the commonsense reasoning, and the symbolic reasoning. The experimental result highlights that our SoftCoT benefits from the supervised fine-tuning and mitigates the catastrophic forgetting problems in state-of-the-art LLMs.

\begin{table*}[t!]
\setlength{\tabcolsep}{4pt}
    \centering
    \begin{tabular}{l| c c c | c | c | c c c c c}
    \toprule
    \multirow{2}{*}{Model} & GSM8K & ASDiv-Aug & AQuA & StrategyQA & DU & Avg. \\
    \cline{2-7}
    ~ & \multicolumn{3}{c|}{Mathematical} & Commonsense & Symbolic \\
    \midrule
    Zero-Shot CoT
    & 83.70$_{\pm \text{0.78}}$ & 87.19$_{\pm \text{0.28}}$ & 64.53$_{\pm \text{3.27}}$ & 49.65$_{\pm \text{3.18}}$ & 66.40$_{\pm \text{2.26}}$ & 70.29 \\
    Zero-Shot CoT-Unk
    & 84.12$_{\pm \text{0.71}}$ & 86.94$_{\pm \text{0.89}}$ & 64.72$_{\pm \text{2.06}}$ & 50.74$_{\pm \text{1.90}}$ & 66.48$_{\pm \text{1.43}}$ & 70.60 \\
    Zero-Shot Assist-CoT
    & 84.85$_{\pm \text{0.35}}$ & 88.63$_{\pm \text{1.05}}$ & 64.96$_{\pm \text{2.83}}$ & 52.71$_{\pm \text{2.65}}$ & 67.04$_{\pm \text{2.84}}$ & 71.64 \\
    \midrule
    LoRA Fine-Tuning
    & 81.80$_{\pm \text{0.00}}$ & 86.80$_{\pm \text{0.00}}$ & 62.60$_{\pm \text{0.00}}$ & - & - & - \\
    Coconut~\cite{DBLP:journals/corr/abs-2412-06769}
    & 82.49$_{\pm \text{0.00}}$ & 86.90$_{\pm \text{0.00}}$ & 63.39$_{\pm \text{0.00}}$ & - & - & - \\
    {\bf SoftCoT (Ours)}
    & {\bf 85.81$_{\pm \text{1.82}}$} & {\bf 88.90$_{\pm \text{1.01}}$} & {\bf 72.44$_{\pm \text{2.19}}$} & {\bf 60.61$_{\pm \text{1.55}}$} & {\bf 67.52$_{\pm \text{2.92}}$} & {\bf 75.06} \\
    \bottomrule
\end{tabular}
\caption{Model performance using Qwen2.5-7B-Instruct. The first three rows are results without training, the last three rows are results trained by language modeling objective. Notably, only the parameters in the projection module is trained for our SoftCoT.
\label{table:softcot-generation-to-qwen2.5}}
\end{table*}

\subsection{Generalization to Other LLM Backbones}

In addition to LLaMA-3.1, we evaluate SoftCoT on another state-of-the-art LLM family: Qwen2.5~\cite{DBLP:journals/corr/abs-2412-15115}. Specifically, we select Qwen2.5-7B-Instruct as the backbone LLM to assess the generalization capability of SoftCoT. As shown in Table~\ref{table:softcot-generation-to-qwen2.5}, our analysis yields the following three key findings:

\noindent (1) {\bf SoftCoT is effective across different backbone models}: Experimental results on Qwen2.5-7B-Instruct show that SoftCoT consistently improves performance across all reasoning datasets, underscoring its robustness. These findings suggest that SoftCoT serves as a generalizable framework that can be effectively adapted to diverse state-of-the-art LLM architectures.

\noindent (2) {\bf SoftCoT enhances LLMs' weaker areas while preserving their strengths}: Experiments on both LLaMA and Qwen LLMs reveal that SoftCoT yields the most significant improvements in commonsense reasoning tasks, where LLMs typically underperform compared to mathematical reasoning. This advantage may stem from SoftCoT's ability to generate contextually relevant continuous thought processes, effectively activating the corresponding knowledge areas within the model. Furthermore, SoftCoT helps mitigate catastrophic forgetting in domains where LLMs already excel, such as mathematical reasoning, thereby preserving and reinforcing existing capabilities.

\noindent (3) {\bf SoftCoT facilitates domain transfer}: Given that the Date Understanding dataset lacks training samples, we train the model on other similar datasets and apply zero-shot transfer to evaluate its generalization on Date Understanding. The results indicate that SoftCoT consistently enhances performance in zero-shot domain transfer scenarios, further demonstrating its adaptability.

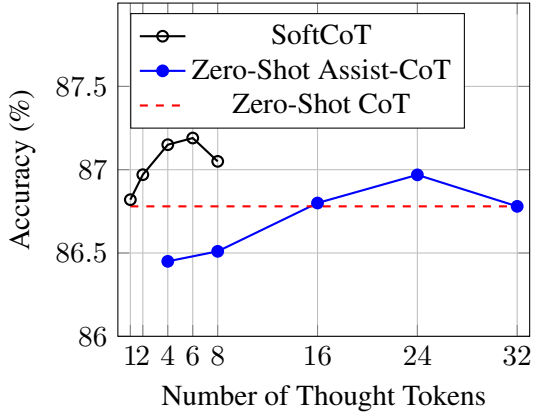
\begin{figure}[!thp]
  \centering
  \begin{tikzpicture}[scale=0.92]
      \begin{axis}[
          width=7cm, height=6cm,
          xlabel={Number of Thought Tokens},
          ylabel={Accuracy (\%)},
          xmin=0, xmax=33,
          ymin=86, ymax=88,
          xtick={1,2,4,6,8,16,24,32},
          ytick={86.0, 86.5, 87.0, 87.5},
          grid=major,
          legend pos=north west,
          every axis plot/.append style={thick}
      ]

      \addplot[color=black, mark=o] coordinates {(1, 86.82) (2, 86.97) (4, 87.15) (6, 87.19) (8, 87.05)};
      \addlegendentry{SoftCoT};

      \addplot[color=blue, mark=*] coordinates {(4, 86.45) (8, 86.51) (16, 86.80) (24, 86.969) (32, 86.78)};
      \addlegendentry{Zero-Shot Assist-CoT};

      \addplot[color=red, dashed] coordinates {(1, 86.78) (32, 86.78)};
      \addlegendentry{Zero-Shot CoT};

      \end{axis}
  \end{tikzpicture}
  \caption{The impact of thought token numbers in ASDiv-Aug using LLaMA-3.1-8B-Instruct.}
  \label{fig:softcot-exp-num-thought-tokens}
\end{figure}

\subsection{Model Analysis and More Studies}

\subsubsection{The Number of Thought Tokens}

To better understand SoftCoT, we conduct experiments to examine the impact of the number of thought tokens. The results, presented in Figure~\ref{fig:softcot-exp-num-thought-tokens}, lead to the following key observations:

\noindent (1) {\bf Soft thoughts reduce the required number of thought tokens}: We observe that SoftCoT achieves optimal performance with only six thought tokens, whereas Zero-Shot Assist-CoT requires 24 thought tokens to reach a similar level of effectiveness. This suggests that soft thoughts, which operate in a continuous space, exhibit a stronger representational capacity than hard thoughts expressed in the discrete language space. Our experiments indicate that the optimal number of hard thought tokens is approximately four times that of soft thought tokens, aligning with the 5x ratio reported by \citet{DBLP:journals/corr/abs-2412-13171}.

\noindent (2) {\bf SoftCoT mitigates the catastrophic forgetting problem}: Experimental results show that SoftCoT consistently outperforms Zero-Shot CoT across all tested numbers of soft thought tokens. In contrast, Zero-Shot Assist-CoT underperforms compared to Zero-Shot CoT when the number of thought tokens is insufficient. This is likely because the assistant model fails to generate a sufficiently informative set of thought tokens under these constraints, introducing noise and leading to confusion in the LLM's reasoning process.

\begin{table*}[t!]
    \centering\small
    \tabcolsep 0.0pt
    \begin{tabular}{l| c c| c c| c c| c c| c c}
    \toprule
    \multirow{2}{*}{Model} & \multicolumn{2}{c|}{GSM8K} & \multicolumn{2}{c|}{ASDiv-Aug} & \multicolumn{2}{c|}{AQuA} & \multicolumn{2}{c|}{StrategyQA} & \multicolumn{2}{c}{DU} \\
    \cline{2-11}
    ~ & $N=1$ & $N=10$ & $N=1$ & $N=10$ & $N=1$ & $N=10$ & $N=1$ & $N=10$ & $N=1$ & $N=10$\\
    \midrule
    Zero-Shot CoT & 79.61$_{\pm \text{0.81}}$ & 90.36$_{\pm \text{0.40}}$ & 86.78$_{\pm \text{0.63}}$ & 89.23$_{\pm \text{0.17}}$ & 54.65$_{\pm \text{2.43}}$ & 63.23$_{\pm \text{0.86}}$ & 65.63$_{\pm \text{3.31}}$ & 70.13$_{\pm \text{0.47}}$ & 54.40$_{\pm \text{2.40}}$ & 65.76$_{\pm \text{1.54}}$ \\
    Zero-Shot Assist-CoT & 80.76$_{\pm \text{1.53}}$ & 90.43$_{\pm \text{0.69}}$ & 86.96$_{\pm \text{0.46}}$ & 89.48$_{\pm \text{0.36}}$ & 55.83$_{\pm \text{2.98}}$ & 63.62$_{\pm \text{0.99}}$ & 66.55$_{\pm \text{3.99}}$ & 70.48$_{\pm \text{0.68}}$ & 58.24$_{\pm \text{3.56}}$ & 65.84$_{\pm \text{1.93}}$ \\
    \midrule
    {\bf SoftCoT (Ours)} & {\bf 81.03$_{\pm \text{0.42}}$} & {\bf 90.63$_{\pm \text{0.39}}$} & {\bf 87.19$_{\pm \text{0.40}}$} & {\bf 89.75$_{\pm \text{0.29}}$} & {\bf 56.30$_{\pm \text{1.67}}$} & {\bf 65.51$_{\pm \text{0.72}}$} & {\bf 69.04$_{\pm \text{1.23}}$} & {\bf 71.14$_{\pm \text{0.10}}$} & {\bf 59.04$_{\pm \text{1.93}}$} & {\bf 67.36$_{\pm \text{1.12}}$} \\
    \bottomrule
\end{tabular}
    \caption{Self Consistency for SoftCoT on LLaMA-3.1-8B-Instruct. ``$N$'' indicates the number of reasoning chains.}\label{table:softcot-experiment-self-consistency}
\end{table*}

\subsubsection{Varying the size of Assistant Model}

\begin{table}[t!]
    \centering\small
    \tabcolsep 4.0pt
    \begin{tabular}{l| c c c}
    \toprule
    Method & 0.5B & 1.5B & 7B \\
    \midrule
    Zero-Shot CoT & 83.70 & 83.70 & 83.70 \\
    Zero-Shot Assist-CoT & 84.78 & 84.85 & 84.90 \\
    \midrule
    SoftCoT & {\bf 85.76} & {\bf 85.81} & {\bf 85.84}\\
    \bottomrule
\end{tabular}
    \caption{Performance on GSM8K with different sizes of assistant model on Qwen2.5 series.}\label{table:softcot-experiment-assistant-model-size}
\end{table}

In the zero-shot assist-CoT setting, the role of the assistant model is to extract or summarize key reasoning cues to guide the target reasoning LLM. While the assistant provides helpful context, the final answer is generated solely by the reasoning LLM. Empirically, we observe that the scale of the assistant model has limited impact on the accuracy of the final answer (see row ``Zero-shot Assist-CoT'' in Table~\ref{table:softcot-experiment-assistant-model-size}).

A similar observation in the SoftCoT setting. Although the assistant model now produces continuous soft thought tokens instead of discrete hard tokens, its fundamental role—providing intermediate reasoning signals—remains unchanged. Our experiments similarly show that varying the assistant model's scale does not significantly affect final task performance (see row ``SoftCoT'' in Table~\ref{table:softcot-experiment-assistant-model-size}).

\subsubsection{Model-Orthogonal Factors}

Self-Consistency~\cite{DBLP:conf/iclr/0002WSLCNCZ23} is a widely adopted technique for enhancing Chain-of-Thought (CoT) reasoning by expanding the search space. One of the most straightforward implementations involves generating multiple CoT reasoning paths and determining the final answer through majority voting. This approach is effective in mitigating errors in reasoning steps by leveraging the diversity of generated thought processes.

To further assess the effectiveness of SoftCoT, we conduct experiments incorporating self-consistency. As shown in Table~\ref{table:softcot-experiment-self-consistency}, SoftCoT consistently outperforms baseline models, demonstrating that its benefits are complementary to those of self-consistency rather than being redundant or conflicting. This suggests that SoftCoT introduces an independent improvement mechanism, which can be effectively combined with self-consistency for enhanced reasoning performance.

A key advantage of SoftCoT in this context is its ability to provide a more expressive and compact representation of intermediate reasoning steps in continuous space. Unlike traditional CoT reasoning, where discrete thought tokens may introduce inconsistencies or redundant reasoning paths, SoftCoT enables more efficient reasoning trajectories with fewer tokens. This allows self-consistency methods to aggregate results from higher-quality reasoning paths, leading to a more robust and accurate final prediction.

\section{Conclusion}

In this paper, we introduce SoftCoT, a soft chain-of-thought prompting approach for efficient LLM reasoning. SoftCoT consists of three steps: (1) an assistant model generates soft thought tokens, (2) a projection module trained to map the soft thoughts to LLM's representation space, and (3) the LLM applies soft thoughts for reasoning. To enhance efficiency, SoftCoT speculatively generates \textit{all} the soft thought tokens in a single forward pass. To mitigate the catastrophic forgetting, SoftCoT freezes the backbone LLM and only tunes the projection module. Experiments on five datasets across three types of reason tasks demonstrate the effectiveness of our proposed SoftCoT. Experiments on multiple LLMs as well as orthogonal method such as self-consistency shows the robustness of SoftCoT, which can be adapted in widely scenarios.

\section*{Acknowledgements}

This research is supported, in part, by the Joint NTU-WeBank Research Centre on Fintech (Award No. NWJ-2020-007), Nanyang Technological University, Singapore. This research is also supported, in part, by the National Research Foundation, Prime Minister’s Office, Singapore under its NRF Investigatorship Programme (NRFI Award No. NRF-NRFI05-2019-0002). Xu Guo thanks the Wallenberg-NTU Presidential Postdoctoral Fellowship. Any opinions, findings and conclusions or recommendations expressed in this material are those of the authors and do not reflect the views of National Research Foundation, Singapore.

\section*{Limitations}
While SoftCoT represents a promising advancement in Chain-of-Thought (CoT) reasoning within a continuous space, several limitations must be acknowledged.

\paragraph{SoftCoT Cannot Fully Replace the Reasoning Path}: Although SoftCoT employs soft thought tokens for reasoning, it does not entirely replace the reasoning path. The decoding stage functions as a search process, which is a crucial component of CoT reasoning. Soft thought tokens primarily serve to enrich the probability space for exploration rather than acting as the search mechanism itself.

\paragraph{Need for Further Empirical Evidence on Scalability}: SoftCoT has been evaluated on LLaMA-3.1-8B-Instruct and Qwen2.5-7B-Instruct. However, larger backbone LLMs exist within the same model families. While its success on models with approximately 7–8 billion parameters suggests potential applicability to larger-scale models, its scalability to extremely large LLMs remains an open question and requires thorough empirical validation.

\bibliography{nlp}
\bibliographystyle{acl_natbib}

\appendix
\section*{Appendix}

\section{Implementation Details}
\label{sec:softcot-appendix-implementation-details}

We use the Huggingface Transformers framework~\cite{wolf-etal-2020-transformers} for training. All models are trained on a single NVIDIA A100-80G GPU. The projection module is trained for 10 epochs and we use the checkpoint after the last epoch for evaluation. We use AdamW~\cite{DBLP:conf/iclr/LoshchilovH19} as the optimizer. The learning rate is set as 1e-5, the weight decay is set as 0.01. To fully utilize the GPU memory, we set the batch size as 8 for Qwen and 16 for LLaMA, which depends on the GPU memory usage. For Zero-Shot CoT-Unk, $L\in \{4, 8, 16, 24\}$ \texttt{[UNK]} tokens are appended and we report the best performance. For Zero-Shot Assist-CoT, we use the assistant model to generate 24 tokens. For SoftCoT, the number of soft thought tokens \textbf{$L$ is set as 4}.

More details can be found in our source code at \url{https://github.com/xuyige/SoftCoT}. To better reproduce our work, we also release a reproduction that includes the training logs, the intermediate checkpoints, and the evaluation logs.

\begin{table*}[t]
\setlength{\tabcolsep}{4pt}
    \centering
    \begin{tabular}{l| c c | c  c | c c}
    \toprule
    \multirow{2}{*}{Model} & \multicolumn{2}{c|}{GSM8K} & \multicolumn{2}{c|}{ASDiv-Aug} & \multicolumn{2}{c}{AQuA} \\
    \cline{2-7}
    ~ & $N=1$ & $N=10$ & $N=1$ & $N=10$ & $N=1$ & $N=10$ \\
    \midrule
    Zero-Shot CoT & 91.86$_{\pm \text{0.41}}$ & 92.22$_{\pm \text{0.47}}$ & 91.70$_{\pm \text{0.26}}$ & 91.97$_{\pm \text{0.13}}$ & 70.00$_{\pm \text{1.56}}$ & 76.77$_{\pm \text{0.62}}$ \\
    Zero-Shot Assist-CoT & 91.90$_{\pm \text{0.50}}$ & 92.68$_{\pm \text{0.17}}$ & 91.64$_{\pm \text{0.16}}$ & 91.91$_{\pm \text{0.28}}$ & 70.16$_{\pm \text{2.35}}$ & 76.77$_{\pm \text{0.79}}$ \\
    \midrule
    Coconut~\cite{DBLP:journals/corr/abs-2412-06769} & 87.95$_{\pm \text{0.00}}$ & 90.37$_{\pm \text{0.00}}$ & 89.40$_{\pm \text{0.00}}$ & 90.37$_{\pm \text{0.00}}$ & 68.50$_{\pm \text{0.00}}$ & 76.38$_{\pm \text{0.00}}$ \\
    {\bf SoftCoT (Ours)} & {\bf 92.48$_{\pm \text{0.29}}$} & {\bf 93.19$_{\pm \text{0.32}}$} & {\bf 91.83$_{\pm \text{0.19}}$} & {\bf 92.14$_{\pm \text{0.15}}$} & {\bf 75.04$_{\pm \text{2.68}}$} & {\bf 80.63$_{\pm \text{1.90}}$} \\
    \bottomrule
\end{tabular}
\caption{Model comparison with baselines on Qwen3-8B~\cite{qwen3} on mathematical reasoning datasets. ``$N$'' indicates the number of reasoning chains. For $N=1$, we do not apply any self-consistency techniques. For $N=10$, we use self-consistency with 10 return sequences to apply multiple reasoning chains. Majority vote is applied to obtain the final prediction answer.
\label{table:softcot-result-qwen3-math}}
\end{table*}
\begin{table*}[t]
\setlength{\tabcolsep}{4pt}
    \centering
    \begin{tabular}{l| c c | c  c }
    \toprule
    \multirow{2}{*}{Model} & \multicolumn{2}{c|}{StrategyQA} & \multicolumn{2}{c}{DU} \\
    \cline{2-5}
    ~ & $N=1$ & $N=10$ & $N=1$ & $N=10$ \\
    \midrule
    Zero-Shot CoT & 69.87$_{\pm \text{0.35}}$ & 70.96$_{\pm \text{0.15}}$ & 80.32$_{\pm \text{2.32}}$ & 84.56$_{\pm \text{0.61}}$ \\
    Zero-Shot Assist-CoT & 69.78$_{\pm \text{0.33}}$ & 70.92$_{\pm \text{0.28}}$ & 80.56$_{\pm \text{1.15}}$ & 84.80$_{\pm \text{1.17}}$ \\
    \midrule
    {\bf SoftCoT (Ours)} & {\bf 70.17$_{\pm \text{0.63}}$} & {\bf 71.18$_{\pm \text{0.15}}$} & {\bf 85.60$_{\pm \text{0.57}}$} & {\bf 87.20$_{\pm \text{0.75}}$} \\
    \bottomrule
\end{tabular}
\caption{Model comparison with baselines on Qwen3-8B~\cite{qwen3} on commonsense and symbolic reasoning datasets. ``DU'' indicates the Date Understanding~\cite{DBLP:journals/tmlr/SrivastavaRRSAF23} dataset. ``$N$'' indicates the number of reasoning chains. For $N=1$, we do not apply any self-consistency techniques. For $N=10$, we use self-consistency with 10 return sequences to apply multiple reasoning chains. Majority vote is applied to obtain the final prediction answer. We run for 5 random seeds and report the average accuracy as well as the standard variance.
\label{table:softcot-result-qwen3-others}}
\end{table*}

\section{Experiments on Qwen3-Series}

In this section, we evaluate the performance of our proposed method on the recently released Qwen3-8B model~\cite{qwen3}, a state-of-the-art large language model that continues the Qwen-series advancements. As presented in Table~\ref{table:softcot-result-qwen3-math} and Table~\ref{table:softcot-result-qwen3-others}, our empirical results reveal several important findings:

\noindent (1) {\bf SoftCoT remains consistently effective when applied to cutting-edge LLMs such as Qwen3-8B}. This demonstrates the generalizability and robustness of the SoftCoT framework across different LLM architectures. Despite architectural differences between Qwen3-8B and prior models such as LLaMA-3.1 or Qwen2.5, the underlying principles of SoftCoT, leveraging soft latent thought representations and their diversity, continue to provide performance gains across diverse reasoning tasks.

\noindent (2) {\bf The key insights obtained from experiments on LLaMA-3.1 and Qwen2.5 seamlessly transfer to Qwen3-8B}. Specifically, we observe that SoftCoT helps alleviate the issue of catastrophic forgetting when scaling up reasoning chains, which is particularly critical when deploying CoT-based strategies in real-world inference scenarios. Additionally, the orthogonality between SoftCoT and self-consistency, previously established in earlier models, holds true for Qwen3-8B. This suggests that the benefits of SoftCoT can be compounded with self-consistency sampling strategies, offering further improvements without redundancy or interference.

Overall, these findings support the claim that SoftCoT is a model-agnostic strategy that effectively adapts to new LLM releases, maintaining its strengths in enhancing reasoning performance and flexibility.

\section{Instruction Templates}
\label{sec:softcot-appendix-templates}

In this section, we release the examples for GSM8K for reference. The instruction template for Zero-Shot CoT is from \citet{DBLP:journals/corr/abs-2409-12183}. We design the task-specific instruction template for the assistant model, which is listed under ``{\bf Input for Assistant Model}'' in the examples of Zero-Shot Assist-CoT and SoftCoT.

In the Zero-Shot Assist-CoT, we require the assistant model to generate 24 tokens to facilitate the reasoning process. In SoftCoT, we use two special token ``<|reserved\_special\_token\_0|>'' and ``<|reserved\_special\_token\_1|>'' to represent the beginning as well as the ending of soft thought tokens, and ``<|end\_of\_text|>'' as the \texttt{[UNK]} token, which is the placeholder for soft thought tokens.

In the following examples, contents under the ``{\bf Answer}'' section is the reasoning steps generated by the LLM, the contents under the ``{\bf Model Prediction}'' section is number that represents the model prediction extracted by pre-defined mannual rules, and the contents under the ``{\bf Evaluate}'' section is the judgement about the model prediction is correct or not.

\clearpage
\onecolumn

\begin{tcolorbox}[
  colback=white, colframe=black, arc=3mm, width=\columnwidth,
  title=\textbf{Example Input-Output for Zero-Shot CoT (GSM8K) with LLaMA-3.1-8B-Instruct},
  coltitle=white, colbacktitle=blue, fonttitle=\bfseries
]
{\bf Input:}

Solve the following math problem efficiently and clearly:

- For simple problems (2 steps or fewer): Provide a concise solution with minimal equation.

- For complex problems (3 steps or more):

Use this step-by-step format:

\#\# Step 1: [Brief calculations]

\#\# Step 2: [Brief calculations]

...

Regardless of the approach, always conclude with:

Therefore, the final answer is: \$ \textbackslash boxed\{\{answer\}\}\$. I hope it is correct.

Where [answer] is just the final number or expression that solves the problem.

Problem: John runs 60 miles a week. He runs 3 days a week.  He runs 3 hours the first day and half as much the other two days he runs.  How fast does he run?

\vspace{5pt}
\textbf{Answer:}

\#\# Step 1: Calculate the total hours John runs in a week

To find out how much he runs in total in a week, we multiply the number of hours he runs on the first day by 3 (for the 3 days in a week) and add the same multiplied by 0.5 (since he runs half as much on the other two days).

\#\# Step 2: Calculate the number of hours John runs in a week

John runs 3 hours on the first day, so he runs (3 * 3) + (3 * 0.5 * 2) hours in a week.

\#\# Step 3: Calculate John's total hours and miles run in a week

3 * 3 = 9 hours for the first day, and 3 * 0.5 * 2 = 3 hours for the other two days. His total hours are 9 + 3 = 12 hours in a week.

\#\# Step 4: Calculate John's speed in miles per hour

John runs 60 miles in 12 hours. So, we divide 60 / 12 to find his speed.

60 / 12 = 5 miles per hour.

Therefore, the final answer is: \$\textbackslash boxed\{5\}\$.

\vspace{5pt}
\textbf{Model Prediction:}

5

\vspace{5pt}
\textbf{Evaluate:}

{\bf \color{red}Incorrect}, the ground-truth is 10 and the model prediction is 5.

\end{tcolorbox}

\begin{tcolorbox}[
  colback=white, colframe=black, arc=3mm, width=\columnwidth,
  title=\textbf{Example Input-Output for Zero-Shot Assist-CoT (GSM8K) with LLaMA-3.1-8B-Instruct},
  coltitle=white, colbacktitle=red, fonttitle=\bfseries
]
{\bf Input for Assistant Model:}

You are required to generate 24 tokens to help another language model to solve the following math reasoning task efficiently and clearly. Here are the requirements of your generated tokens:

- The tokens should include some useful information for the reasoning problem, for example, the numbers and the operations needed for calculation.

- Generate the tokens starts from the most important or the highest related tokens.

- **Informative tokens are required**: (1) Do not need to generate a sentence or paragraph, (2) Do not need to generate the uninformative tokens such as serial number.

- The tokens should be useful for large language model to answer the question with the numbers.

...

Here is the problem: John runs 60 miles a week. He runs 3 days a week. He runs 3 hours the first day and half as much the other two days he runs.  How fast does he run?

\vspace{5pt}

{\bf Input for Large Reasoning Model:}

Solve the following math problem efficiently and clearly:

- For simple problems (2 steps or fewer): Provide a concise solution with minimal equation.

- For complex problems (3 steps or more):

Use this step-by-step format:

\#\# Step 1: [Brief calculations]

\#\# Step 2: [Brief calculations]

...

Regardless of the approach, always conclude with:

Therefore, the final answer is: \$ \textbackslash boxed\{\{answer\}\}\$. I hope it is correct.

Where [answer] is just the final number or expression that solves the problem.

Problem: John runs 60 miles a week. He runs 3 days a week.  He runs 3 hours the first day and half as much the other two days he runs.  How fast does he run?

There are some prompts generated by a weaker assistant model. Some prompts maybe useful while others maybe unuseful for your reasoning. If the prompts are correct, you can use it as reference. If the prompts are not correct, you can ignore them and focus back to solving the problem.

Here are prompts:

1. a

2. miles

3. week

4. days

5. 3-mile

6. run

\vspace{5pt}
\textbf{Answer:}

\#\# Step 1: Calculate the total hours John runs in a week

John runs 3 hours on the first day. On the other two days, he runs half as much, which is 3 / 2 = 1.5 hours per day. So, the total hours John runs in a week is 3 + 1.5 + 1.5 = 6 hours.

\#\# Step 2: Calculate John's speed

To find John's speed, divide the total distance he runs in a week (60 miles) by the total hours he runs in a week (6 hours). Speed = Distance / Time = 60 miles / 6 hours = 10 miles per hour.

Therefore, the final answer is: \$\textbackslash boxed\{10\}\$.

\vspace{5pt}
\textbf{Model Prediction:}

10

\vspace{5pt}
\textbf{Evaluate:}

{\bf \color{darkgreen}Correct!}

\end{tcolorbox}

\newpage

\begin{tcolorbox}[
  colback=white, colframe=black, arc=3mm, width=\columnwidth,
  title=\textbf{Example Input-Output for SoftCoT (GSM8K) with LLaMA-3.1-8B-Instruct},
  coltitle=white, colbacktitle=gray, fonttitle=\bfseries
]
{\bf Input for Assistant Model:}

<user>:

You are required to generate 4 tokens to help another language model to solve the following math reasoning task efficiently and clearly. Here are the requirements of your generated tokens:

- The tokens should include some useful information for the reasoning problem, for example, the numbers and the operations needed for calculation.

- Generate the tokens starts from the most important or the highest related tokens.

- **Informative tokens are required**: (1) Do not need to generate a sentence or paragraph, (2) Do not need to generate the uninformative tokens such as serial number.

- The tokens should be useful for large language model to answer the question with the numbers.

...

Here is the problem: John runs 60 miles a week. He runs 3 days a week.  He runs 3 hours the first day and half as much the other two days he runs.  How fast does he run?

<assistant>:

Here are 4 tokens to help the language model solve this math reasoning task: <|reserved\_special\_token\_0|> <|end\_of\_text|> <|end\_of\_text|> <|end\_of\_text|> <|end\_of\_text|> <|reserved\_special\_token\_1|>

\vspace{5pt}

{\bf Input for Large Reasoning Model:}

Solve the following math problem efficiently and clearly:

- For simple problems (2 steps or fewer): Provide a concise solution with minimal equation.

- For complex problems (3 steps or more):

Use this step-by-step format:

\#\# Step 1: [Brief calculations]

\#\# Step 2: [Brief calculations]

...

Regardless of the approach, always conclude with:

Therefore, the final answer is: \$ \textbackslash boxed\{\{answer\}\}\$. I hope it is correct.

Where [answer] is just the final number or expression that solves the problem.

Problem: John runs 60 miles a week. He runs 3 days a week.  He runs 3 hours the first day and half as much the other two days he runs.  How fast does he run?

There are some prompts generated by a weaker assistant model. Some prompts maybe useful while others maybe unuseful for your reasoning. If the prompts are correct, you can use it as reference. If the prompts are not correct, you can ignore them and focus back to solving the problem.

Here are prompts: <|reserved\_special\_token\_0|> <|end\_of\_text|> <|end\_of\_text|> <|end\_of\_text|> <|end\_of\_text|> <|reserved\_special\_token\_1|>

\vspace{5pt}
\textbf{Answer:}

\#\# Step 1: Calculate the total time John runs each week in hours. John runs 3 hours on the first day. For the other two days, he runs half as much, which is 3 / 2 = 1.5 hours per day. So, his total running time each week is 3 + 2 * 1.5 = 6 hours.

\#\# Step 2: Determine John's speed in miles per hour by dividing his total weekly distance by his total weekly running time. John runs 60 miles a week and his total running time each week is 6 hours. Thus, his speed is 60 / 6 = 10 miles per hour.

Therefore, the final answer is: \$\textbackslash boxed\{10\}\$.

\vspace{5pt}
\textbf{Model Prediction:}

10

\vspace{5pt}
\textbf{Evaluate:}

{\bf \color{darkgreen}Correct!}

\end{tcolorbox}


\end{document}